%% file: 00-main.tex
\documentclass[letterpaper]{article} 
\usepackage[]{aaai25}  

\usepackage{times}  
\usepackage{helvet}  
\usepackage{courier}  
\usepackage[hyphens]{url}  
\usepackage{graphicx} 
\usepackage{natbib}  
\usepackage{caption} 
\usepackage{algorithm}
\usepackage{algorithmic}
\usepackage{tikz}
\usepackage{amssymb}

\usepackage{newfloat}
\usepackage{listings}
\usepackage{subfig}
\usepackage{graphicx}

\DeclareCaptionStyle{ruled}{labelfont=normalfont,labelsep=colon,strut=off} 
\floatstyle{ruled}
\newfloat{listing}{tb}{lst}{}
\floatname{listing}{Listing}
\title{PrecipDiff: Leveraging image diffusion models to enhance satellite-based precipitation observations}
\author{
    Ting-Yu Dai \textsuperscript{\rm 1 2}, 
    Hayato Ushijima-Mwesigwa \textsuperscript{\rm 2}
}
\affiliations{
        \textsuperscript{\rm 1} The University of Texas at Austin\\
    \textsuperscript{\rm 2} Fujitsu Research of America, Inc.\\
    
       funnyengineer@utexas.edu, hayato@fujitsu.com
}

\usepackage{bibentry}

\begin{document}

\maketitle
\input{1-abstract}
\input{2-introduction}
\input{3-background}
\input{4-methodology}
\input{5-data}
\input{6-results}

\input{7-discussion}
\input{8-conclusion}
\clearpage 

\bibliography{aaai25}

\end{document}

%% file: 1-abstract.tex
\begin{abstract}

A recent report from the World Meteorological Organization (WMO) highlights that water-related disasters have caused the highest human losses among natural disasters over the past 50 years, with over 91\% of deaths occurring in low-income countries. This disparity is largely due to the lack of adequate ground monitoring stations, such as weather surveillance radars (WSR), which are expensive to install. For example, while the US and Europe combined possess over 600 WSRs, Africa, despite having almost one and half times their landmass, has fewer than 40.
To address this issue, satellite-based observations offer a global, near-real-time monitoring solution. However, they face several challenges like accuracy, bias, and low spatial resolution. 
This study leverages the power of diffusion models and residual learning to address these limitations in a unified framework. We introduce the first diffusion model for correcting the inconsistency between different precipitation products. Our method demonstrates the effectiveness in downscaling satellite precipitation estimates from 10 km to 1 km resolution. Extensive experiments conducted in the Seattle region demonstrate significant improvements in accuracy, bias reduction, and spatial detail. Importantly, our approach achieves these results using only precipitation data, showcasing the potential of a purely computer vision-based approach for enhancing satellite precipitation products and paving the way for further advancements in this domain.

\end{abstract}

%% file: 2-introduction.tex
\section{Introduction}

Water-related disasters, such as landslides, floods, and droughts, constitute a significant majority of natural disasters. The past five decades have witnessed over 11,000 reported weather-related disasters globally, tragically claiming over two million lives and inflicting \$3.64 trillion in economic losses \cite{wmo2021weather}. Developing countries disproportionately bear this burden, accounting for over 91\% of these fatalities. However, advancements in early warning systems and disaster management have yielded a nearly threefold reduction in fatalities between 1970 and 2019, underscoring the life-saving potential of technological progress.


Precipitation monitoring forms a cornerstone of effective early warning systems, proving crucial for forecasting and mitigating weather-related hazards. Weather surveillance radar (WSR) systems offer high-resolution data on precipitation intensity and distribution, invaluable for short-term forecasting and nowcasting. However, the prohibitive cost of WSRs, often reaching millions of dollars per unit, leads to uneven global distribution. For example, the U.S. and Europe, with a combined population of roughly 1 billion, possess nearly 700 WSRs. In stark contrast, Africa, with a larger population and a landmass more than 1.5 times, operates fewer than 40 WSRs \cite{aljazeera2023africa}.

Satellite-based precipitation products (SPPs) offer a valuable alternative, providing continuous spatiotemporal estimation with global coverage. Products such as the Tropical Rainfall Measuring Mission (TRMM), the Climate Prediction Center Morphing Technique (CMORPH) \cite{joyce2004cmorph}, the Global Precipitation Measurement (GPM) mission \cite{hou2014global}, and Global Satellite Mapping of Precipitation (GSMap) \cite{kubota2020global}, along with advancements like the Integrated Multi-satellite Retrievals for GPM (IMERG), have significantly improved precipitation monitoring capabilities. 

However, SPPs still face challenges. These products often exhibit biases compared to ground-based observations due to various factors, including limitations in sensor technology, retrieval algorithms, cloud cover, and the inherent complexities of precipitation processes. Their indirect estimation nature, coupled with limitations in spatiotemporal sampling, instrument capabilities, and retrieval algorithms, can lead to biases and relatively low accuracy. Comprehensive evaluation and bias correction are therefore crucial, especially for early warning systems \cite{yin2008assessment,ji2012characterizing,yang2014using,lu2018evaluation,lu2018evaluationb}. Furthermore, the coarse spatial resolution of current SPPs (typically 0.1° or lower) limits their applicability in urban settings where finer resolutions are required \cite{berne2004temporal}.

This work addresses these two key challenges of SPPs: bias correction and spatial resolution enhancement through downscaling. 
The main contributions of this paper are:

\begin{enumerate}
\item We formulate both bias correction and downscaling as a residual learning problem and propose using diffusion models to solve them in a unified framework.
\item We introduce the first computer vision-based bias correction algorithm, which uses the diffusion model to learn the biases of SPPs.
\item We demonstrate the practical application of diffusion models for operational datasets. Moving a step beyond previous related work, that primarily focused on simulated data, we showcase their effectiveness in real-time downscaling tasks using state-of-the-art deep learning technique
\item Extensive experiments demonstrate the effectiveness of our proposed method in both bias correction and downscaling tasks, even without incorporating other related variables such as elevation, temperature, humidity, etc, which is the common practice for tackling these problems. 
\end{enumerate}

%% file: 3-background.tex
\section{Background}

\begin{figure*}[t]
    \centering
   \includegraphics[width=\linewidth]{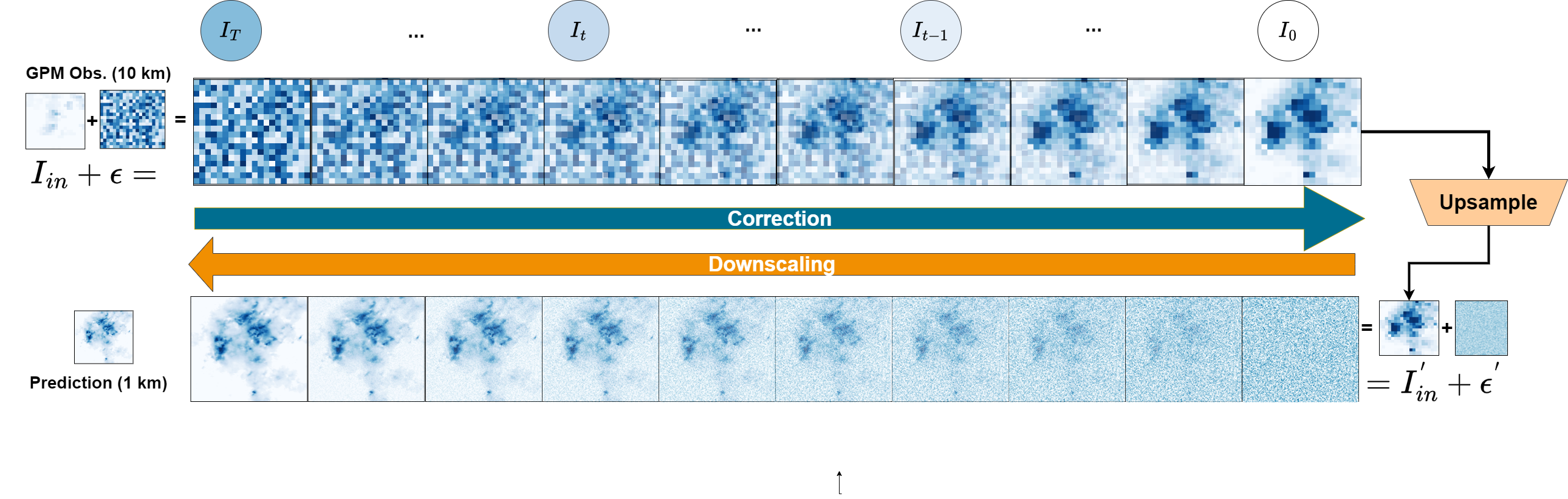}
    \caption{Overview flowchart for inference process. Satellite precipitation data are correctized by the first diffusion process and then upsampled using a linear method. Second diffusion process are applied to synthesized downscaled predictions to create high-resolution data.}
    \label{fig:inference}
\end{figure*}


\subsection{Bias Correction in SPPs}

Accurate bias correction is essential for enhancing the reliability of SPPs and ensure their effective application in diverse domains, including hydrological modeling, climate monitoring, and disaster preparedness.

Numerous methods have been proposed for correcting biases in SPPs, broadly categorized into statistical, machine learning, and hybrid approaches. Traditional statistical techniques encompass linear regression \cite{wilks2011statistical}, quantile mapping \cite{scheuerer2014multivariate}, multiplicative bias correction \cite{wilks2011statistical}, and power transformation \cite{wilks2011statistical}. Machine learning methods, on the other hand, offer the flexibility to capture nonlinear relationships between SPP estimates and rain gauge data, potentially leading to more accurate corrections. Examples of machine learning approaches include Support Vector Machines, random forests, and deep learning techniques \cite{chen2019deep}. Despite significant progress in bias correction methodologies, it remains an active area of research within the field of SPP development and application.


\subsection{Challenges in Precipitation Downscaling}

Precipitation downscaling presents unique challenges compared to downscaling other meteorological variables like temperature or wind. Its highly variable nature in both space and time, influenced by factors such as orography, land-sea contrasts, and convective processes, makes capturing fine-scale precipitation patterns difficult, particularly in regions with complex terrain or limited observational networks.

Furthermore, extreme precipitation events, crucial for flood risk assessments and water resource management, are often poorly represented in downscaled datasets. This is primarily due to their rarity and the inherent limitations of both dynamical and statistical downscaling methods in capturing these infrequent high-intensity events \cite{hewitson2014interrogating}. While temperature and other variables typically exhibit smoother spatial patterns amenable to downscaling, precipitation often involves abrupt changes over short distances, further complicating the modeling process. These challenges underscore the need for more sophisticated downscaling techniques that can effectively account for the complexities of precipitation processes and provide more reliable, high-resolution estimates for climate impact assessments.

\subsection{Diffusion Models}
Diffusion models \cite{ho2020denoising,sohl2015deep} have emerged as a powerful class of generative models, demonstrating remarkable success in image synthesis, surpassing even the quality of GANs in certain applications. Their applicability extends to both unconditional \cite{ho2020denoising,nichol2021improved,song2020score} and conditional \cite{ho2022cascaded,nichol2021glide,nichol2021improved,preechakul2022diffusion,ramesh2022hierarchical,rombach2022high} generation settings. Beyond image synthesis, diffusion models have rapidly found utility in diverse domains, including audio and video generation, image segmentation, language translation, and recently, climate downscaling.

A denoising diffusion probabilistic model (DDPM) \cite{} employs two Markov chains: a forward process that gradually injects noise into the data and a reverse process that learns to progressively remove noise to recover the original data distribution. Formally, given a data distribution $x_0 \sim q(x_0)$, the forward Markov process generates a sequence of random variables $x_1, \dots, x_T$ with transition kernel $q(x_t|x_{t-1})$. Leveraging the Markov property, the joint distribution is given by:

\begin{equation}
q(x_1, \dots, x_T) = \prod_{t=1}^Tq(x_t|x_{t-1}).
\end{equation}

In DDPMs, the transition kernel is commonly defined as:

\begin{equation}
q(x_t|x_{t-1})= \mathcal{N}(x_t;\sqrt{1-\beta_t}x_{t-1}, \beta_t I ),
\end{equation}
where $\beta_t \in (0,1)$ is a hyperparameter controlling the variance schedule. This forward process gradually adds noise to the original data until it becomes essentially pure noise.

To generate new data samples, a noise vector is first sampled from the prior distribution. The reverse process, parameterized by a prior distribution $p(x_T) = \mathcal{N}(x_T;\textbf{0}, \textbf{I})$ and a learnable transition kernel $p_{\theta}(x_{t-1}|x_t)$, then gradually removes the noise. This kernel is defined as:

\begin{equation}
p_{\theta}(x_{t-1}|x_t) = \mathcal{N}(x_{t-1};\mu_{\theta}(x_t,t), \Sigma_{\theta}(x_t,t))
\end{equation}
where $\theta$ represents the model parameters, and the mean $\mu_{\theta}(x_t,t)$ and variance $\Sigma_{\theta}(x_t, t)$ are parameterized by deep neural networks.

%% file: 4-methodology.tex
\section{Methodology}

\subsection{Residual-Based Diffusion Model Training}
Inspired by Corrdiff \cite{mardani2024residual}, we shift our focus from directly generating high-resolution (HR) precipitation data to learning the residual difference between low-resolution (LR) and HR data. Direct HR generation using diffusion models often results in a loss of intensity and detail in cloud formations and weather fronts. Instead, we train our model to generate the residual, conditioned on the LR data, using the EDM scheduler and sampler. This residual is then added to the interpolated LR image to obtain the downscaled precipitation data. A classic U-Net architecture is employed for denoising the noisy residual based on the EDM noise scheduling algorithm. A key assumption in this residual training approach is the availability of high-quality LR data as a foundation.

\subsection{Training and Inference Pipeline}
\begin{figure*}[t]
\centering
\includegraphics[width=0.8\linewidth]{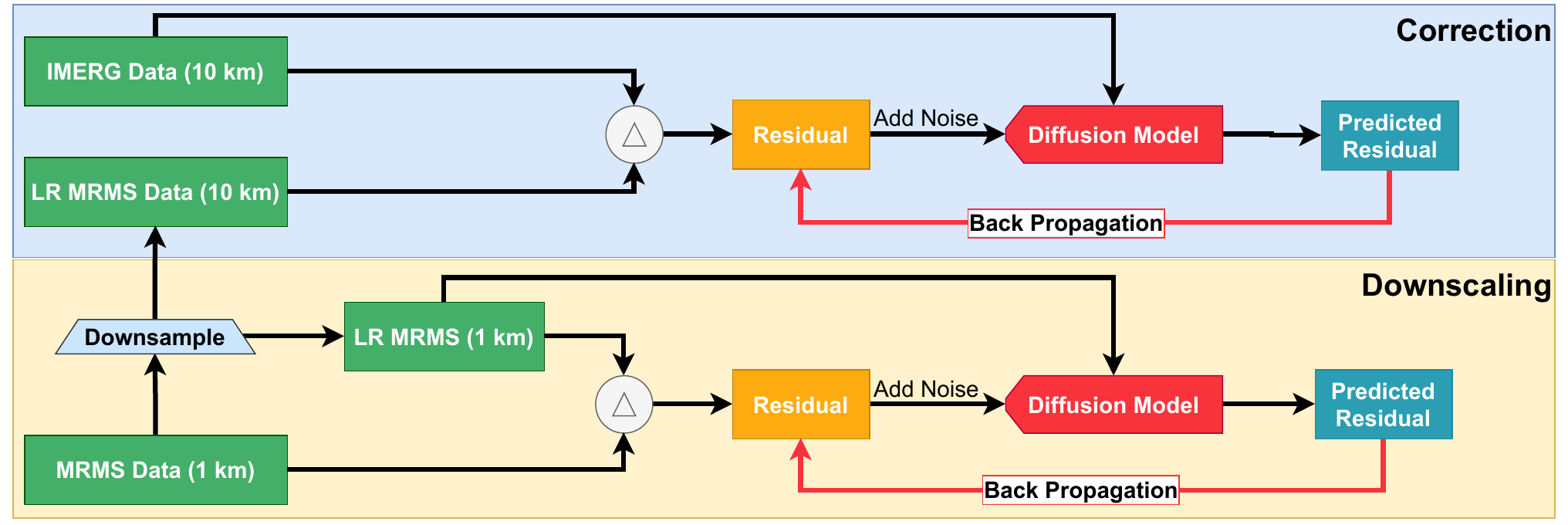}
\caption{Overview flowchart for training correction and downscaling diffusion models. The correction model is trained at 10 km resolution using noisy residuals as input and IMERG data as conditioning information. Residuals at this stage are calculated by subtracting IMERG data from LR MRMS data. The downscaling model is trained at 1 km resolution, with residuals computed by subtracting interpolated LR MRMS data from original MRMS data.}
\label{fig:training}
\end{figure*}

Figures \ref{fig:inference} and \ref{fig:training} illustrate the separate processes of training and inference, respectively. Our framework addresses two primary tasks: bias correction and downscaling.

\subsubsection{Bias Correction}
This task aims to mitigate discrepancies between satellite-based GPM and radar observations, which arise from differences in measurement techniques and processing algorithms. As shown in Figure \ref{fig:training}, the correction model is trained on residuals calculated by subtracting IMERG data from LR MRMS data. IMERG data is used as conditioning information to guide the denoising of noisy residuals.

\subsubsection{Downscaling}
For downscaling, MRMS data is first downsampled to the same resolution as satellite observations, creating an LR representation. Residuals are then calculated by comparing the original MRMS data with the downsampled LR MRMS data, representing the information lost during downsampling. The same residual diffusion learning process is applied to this task, with the coarsened radar observations serving as conditioning information.

Figure \ref{fig:inference} depicts the operational workflow. First, satellite GPM observations are calibrated using the trained bias correction diffusion model. The corrected data is then upsampled to 1 km resolution using linear interpolation and fed as conditioning information to the downscaling diffusion model. The model generates the residual, which is added to the corrected and upsampled data to produce the final high-resolution precipitation prediction. This pipeline enables the generation of corrected and downscaled satellite precipitation data in real-time, with accuracy comparable to radar products and at the same spatial resolution.

\subsection{Bias Correction Model}
The diffusion model for bias correction learns to synthesize the distribution difference between LR satellite estimates and coarsened HR radar data. This addresses the inherent bias between satellite and radar observations due to their different measurement principles and locations. The model effectively corrects this systematic bias, aligning satellite observations more closely with radar observations. To achieve this, radar observations are first coarsened to the same resolution as satellite data using maximum value interpolation to preserve precipitation intensity. Residual training is then applied using the coarsened radar data and the corresponding satellite precipitation data.

\subsection{Downscaling Model}

After bias correction, the same residual training strategy is employed for downscaling. The diffusion model is trained on the residual between coarsened radar precipitation and the original high-resolution radar data, aiming to capture fine-scale details present in the radar data. Coarsened radar observations serve as conditioning information, guiding the model to generate the denoised residual. Importantly, the downscaling model operates independently of the bias correction stage, as the corrected satellite observations may still exhibit differences compared to LR radar images. By training solely on radar data, we minimize uncertainties stemming from input data discrepancies, enabling the model to effectively enhance the spatial resolution of corrected satellite precipitation estimates.

%% file: 5-data.tex
\section{Experiments}

\subsection{Data}

To closely mimic real-world scenarios, we utilize two operational precipitation datasets: the IMERG dataset as our LR satellite observations and the MRMS dataset as our HR radar observations.

\subsubsection{IMERG}

The Integrated Multi-satellitE Retrievals for the Global Precipitation Measurement (GPM) mission (IMERG) is a product of the U.S. GPM Science Team, providing globally covered precipitation estimates with approximately 10 km resolution based on the GPM satellite constellation. IMERG offers three product qualities: early, late, and final run, reflecting data latency and quality. This study utilizes the final run product from version 07.

\subsubsection{MRMS}
The Multi-Radar Multi-Sensor (MRMS) system is a sophisticated operational hydrometeorological framework. It integrates observations and information from various radar networks, satellites, and surface observational systems to generate precipitation estimates at 1 km resolution. MRMS offers various temporal frequencies (daily to 2-minute) and data quality levels, including products fused with ground rain gauge data. To align with the temporal scale of IMERG and leverage the most comprehensive ground information, we utilize the hourly Q3 Multi-sensor Pass 2 precipitation product.

\subsection{Experiment Setup}

Our study focuses on the Seattle region, characterized by frequent rain events. We use data from both IMERG and MRMS for Year 2022 and 2023. To focus on rain events, we filter out data samples with a low proportion of rain, specifically those where the proportion of zero precipitation values within the selected region is below 0.5. This results in approximately 1600 data samples. We then randomly split the dataset into training (90\%) and testing (10\%) sets.
To evaluate the effectiveness of our diffusion model approach for both correction and downscaling tasks independently, we design three experiments: correction task, downscaling task, and the unified correction and downscaling task.

\subsubsection{Unified correction and downscaling task}  This experiment simulates a real-world application scenario. First, we correct the IMERG data using the trained correction model. Then, we use this corrected IMERG data as input to the trained downscaling model. Finally, we compare the resulting downscaled IMERG product with the original HR MRMS data to evaluate the overall effectiveness of our framework.

\subsubsection{Correction task} The target is the residual between the original IMERG data and the coarsened (LR) MRMS data. The input (conditioning image) is the original IMERG data.
\subsubsection{Downscaling task} The target is the residual between the LR MRMS data and the original (HR) MRMS data. The input (conditioning image) is the LR MRMS data.
\subsubsection{Model Configurations and Training}
The correction model is trained on 20x20 pixel image patches using a 3-block U-Net with latent channels of (32, 64, 128) while the downscaling model is rained on 184x200 pixel images (corresponding to the 1 km resolution target) using a 4-block U-Net with latent channels of (128, 256, 256, 512).
Both models are trained for 2000 epochs with a learning rate of 2e-4 and a batch size of 16, distributed across two Quadro RTX 8000 GPUs. Total training time is approximately 10 hours.

\subsubsection{Baseline Model}
We compare our correction model with a supervised Swin2SR \cite{conde2022swin2sr} model to assess the effectiveness of our approach against a state-of-the-art supervised method. The Swin2SR model is trained with the same configuration and a comparable number of trainable parameters as the correction diffusion model, using original IMERG data as input and LR MRMS data as the target.

\subsubsection{Inference}
During inference, we employ the second-order EDM stochastic sampler proposed by \cite{karras2022elucidating} to sample the residual. The correction model first generates a residual, which is added to the original IMERG data to produce the corrected IMERG product. This corrected product then serves as input to the downscaling model, which generates a second residual. Adding this residual to the corrected IMERG data yields the final downscaled IMERG product. Both models utilize 25 steps to solve the reverse SDE process during inference.

%% file: 6-results.tex
\section{Results \& Discussions}

\begin{figure}[t]
    \centering
    \includegraphics[width=\linewidth]{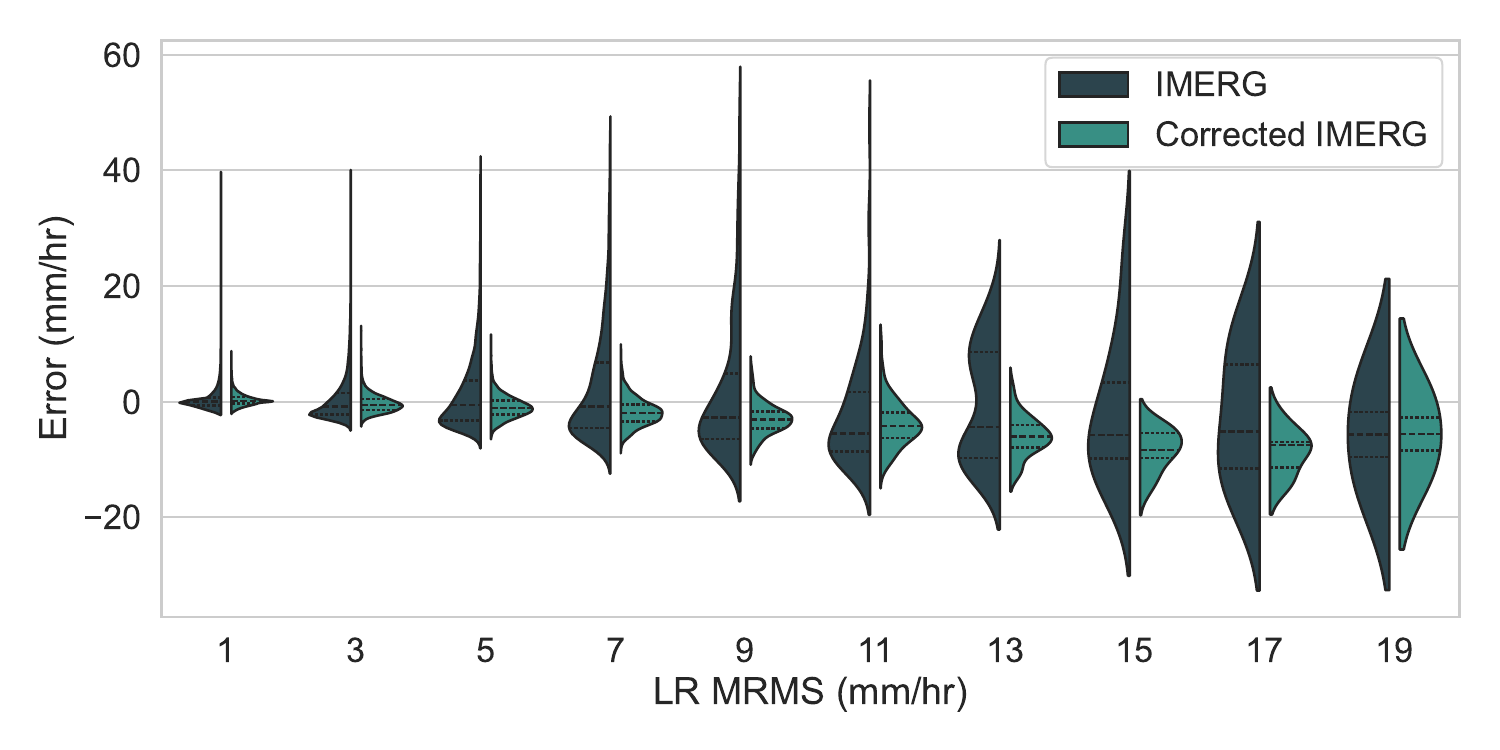}
    \caption{A comparison of the error distribution between original IMERG and Corrected IMERG. The errors are calculated based on the LR MRMS observations, with all pixels flattened for the comparison.}
    \label{fig:vio-I2LRM}
\end{figure}

\begin{table}[t]
    \centering
    \begin{tabular}{c | c c c c }
     \hline
        Method & RMSE & CRPS & CC & SSIM \\ \hline
        Bicubic LR& 3.7087 & 1.9914 & 0.4593 & \textbf{0.5222} \\ 
        Swin2SR & 2.2842 & 1.8493  & \textbf{0.5685} & 0.3418 \\
        PrecipDiff (Ours) & \textbf{1.7969} & \textbf{1.2464}  & 0.4961 & 0.3772 \\  \hline
    \end{tabular}
    \caption{Evaluation for correction task: \textbf{IMERG to LR MRMS}. RMSE, CRPS, Correlation coeiffient, and SSIM assess the overall performance and image synthesis quality between LR MRMS data. The denoted bold values indicate the best performance.}
    \label{tab:correction}
\end{table}

We evaluate our models using four metrics:
\begin{itemize}
    \item Root Mean Squared Error (RMSE): Measures overall performance and the magnitude of errors.
\item Continuous Ranked Probability Score (CRPS): Equivalent to Mean Absolute Error (MAE) for deterministic predictions, assessing overall accuracy. Comparing RMSE and CRPS helps determine if the model primarily improves extreme values (lower RMSE) or general pixel values.
\item Pearson Correlation Coefficient (CC): Evaluates the linear relationship between predictions and ground truth, assessing the model's ability to capture overall precipitation patterns.
\item Structural Similarity Index Measure (SSIM): Measures the similarity of image structures, assessing the model's ability to reproduce spatial details.
\end{itemize}

\subsection{Bias Correction}

The goal of bias correction is to calibrate satellite observations to a scale comparable with ground-based radar data. Directly applying a diffusion model from IMERG to MRMS proved challenging due to consistent biases between these operational datasets. Therefore, we focused on predicting LR MRMS data from IMERG data.
Initially, we experimented with a Swin2SR model \cite{mardani2024residual} for bias correction. While this supervised approach yielded promising metrics, the predictions were overly smooth, resulting in the loss of distinct storm shapes.
To address this, we employed our proposed residual diffusion model for bias correction. As shown in Table \ref{tab:correction}, our model outperforms both the Swin2SR model and bicubic linear interpolation in RMSE and CRPS. One possible explanation for the weaker performance relative to Swin2SR in CC and Bicubic LR in SSIM is that our approach incorporates the sampled residual into the initial Low-Res images. This integration might disturb the overall image structure, causing to lower scores in these two metrics that assess overall similarity.

Figure \ref{fig:vio-I2LRM} compares the error distributions of the original and corrected IMERG data relative to LR MRMS data. IMERG tends to overestimate lower precipitation values and underestimate higher values. Our corrected predictions effectively mitigate the overestimation issue and reduce variation in higher values, although improvements in underestimation are limited.
We also experimented with correcting IMERG data at high resolution by interpolating IMERG and using the residuals between the interpolated data and HR MRMS for training. However, this approach yielded suboptimal results due to the abundance of zero-value pixels in HR MRMS data, which led the model to produce unrealistically low precipitation values.

\begin{figure}[t]
    \centering
    \includegraphics[width=\linewidth]{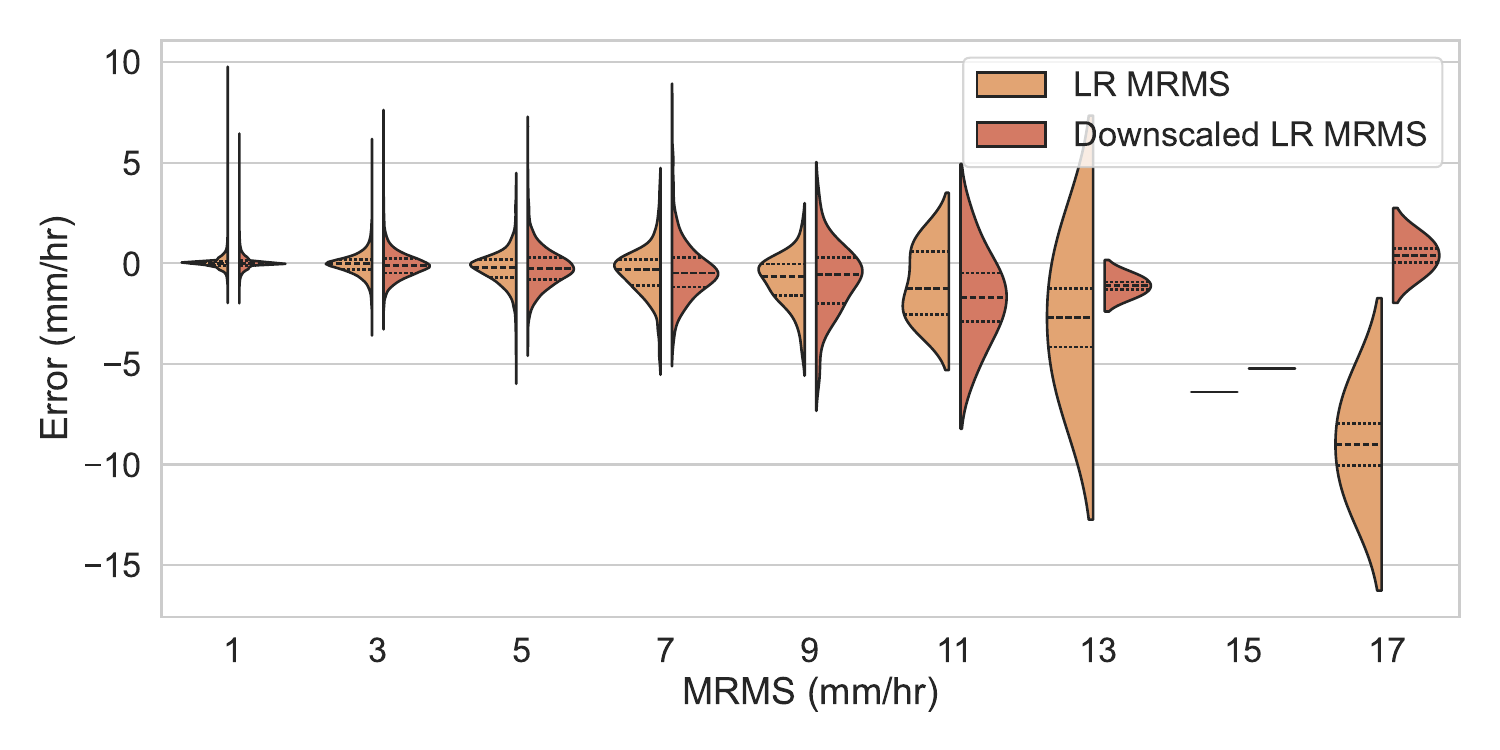}
    \caption{A comparison of the error distributions between LR MRMS and downscaled MRMS. The errors are calculated based on the original MRMS observations, with sampling from the pixels flattened for the comparison.}
    \label{fig:vio-M2M}
\end{figure}

\begin{table}[t]
    \centering
    \begin{tabular}{c|cccc}
     \hline
        Method & RMSE & CRPS & CC & SSIM \\ \hline
        LinearRegression & 0.8642 & 0.5999 & 0.9347 & 0.7378 \\
        PrecipDiff (Ours) & \textbf{0.4543} &\textbf{ 0.2685}  & \textbf{0.9386} &\textbf{ 0.8249} \\ \hline
    \end{tabular}
    \caption{Evaluation for downscaling task: \textbf{LR MRMS to MRMS}. RMSE, CRPS, Correlation coeiffient, and SSIM are used to evaluate the overall performance and image synthesis quality between original MRMS measurements.  The denoted bold values indicate the best performance.}
    \label{tab:downscaling}
\end{table}
\begin{figure*}[t]
    \centering
    \subfloat[\centering IMERG]{{\includegraphics[width=.24\linewidth]{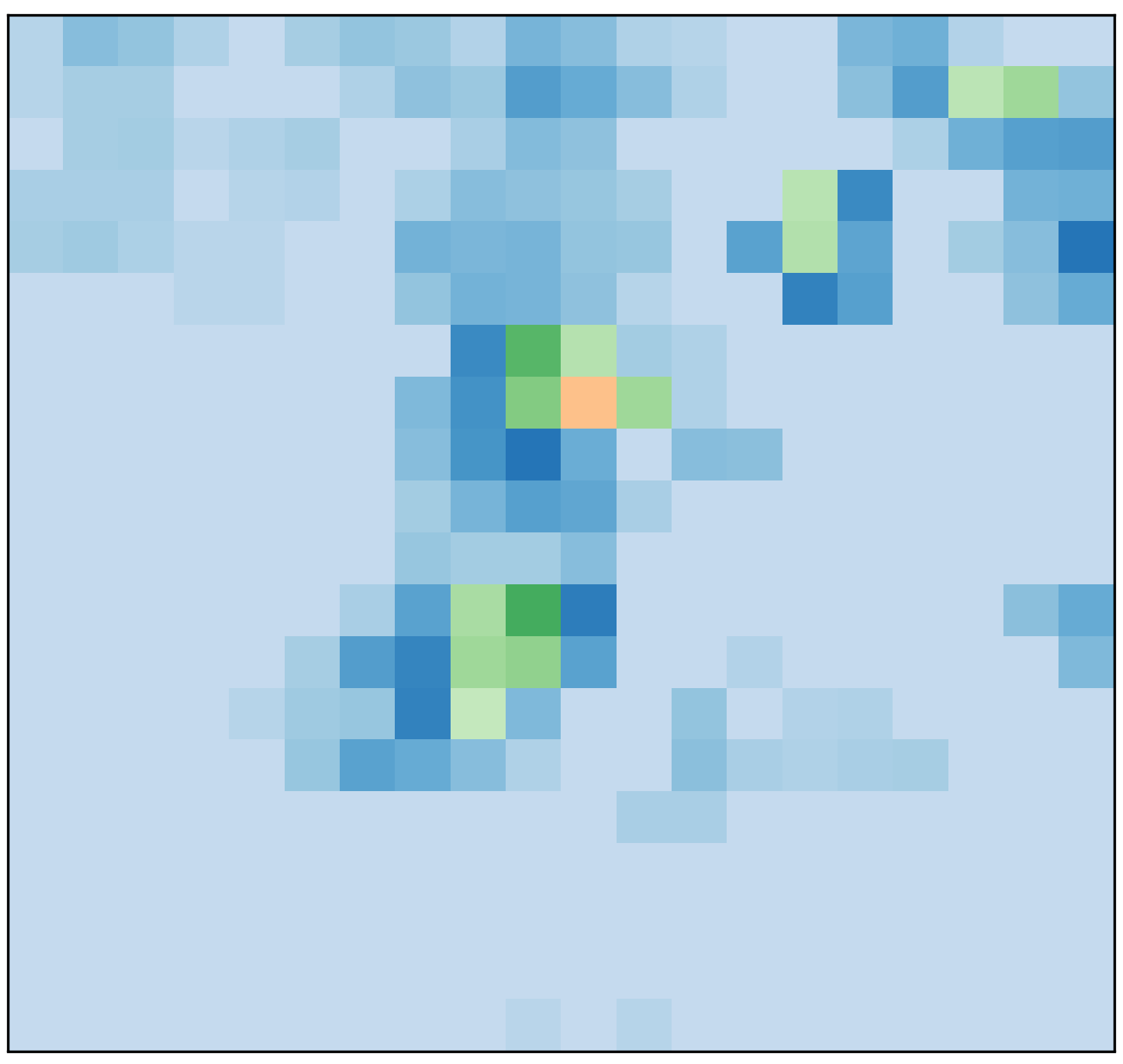}} }
    \subfloat[\centering Corrected IMERG]{{\includegraphics[width=.24\linewidth]{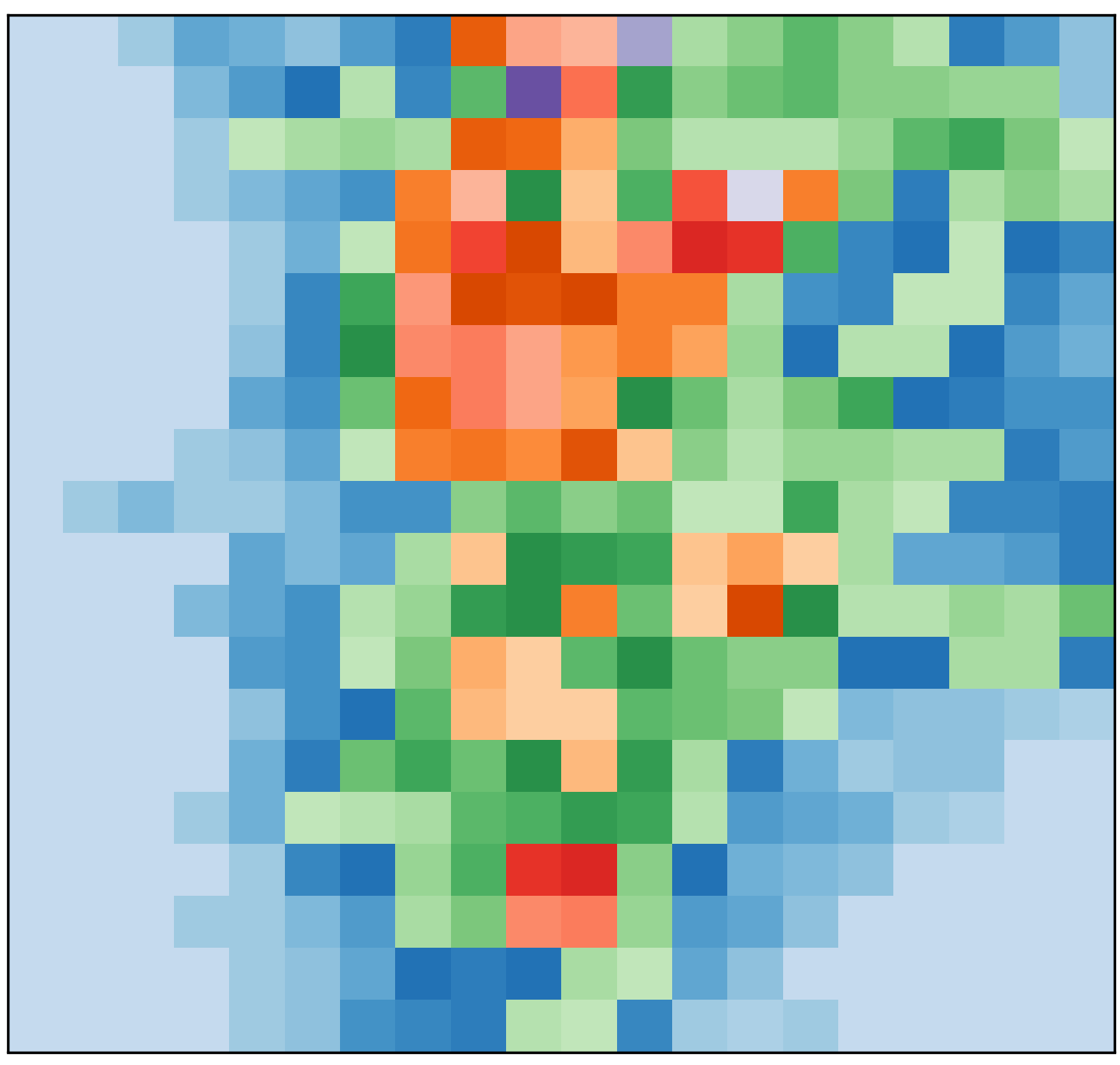} }}
    \subfloat[\centering Corrected and Downscaled IMERG]{{\includegraphics[width=.245\linewidth]{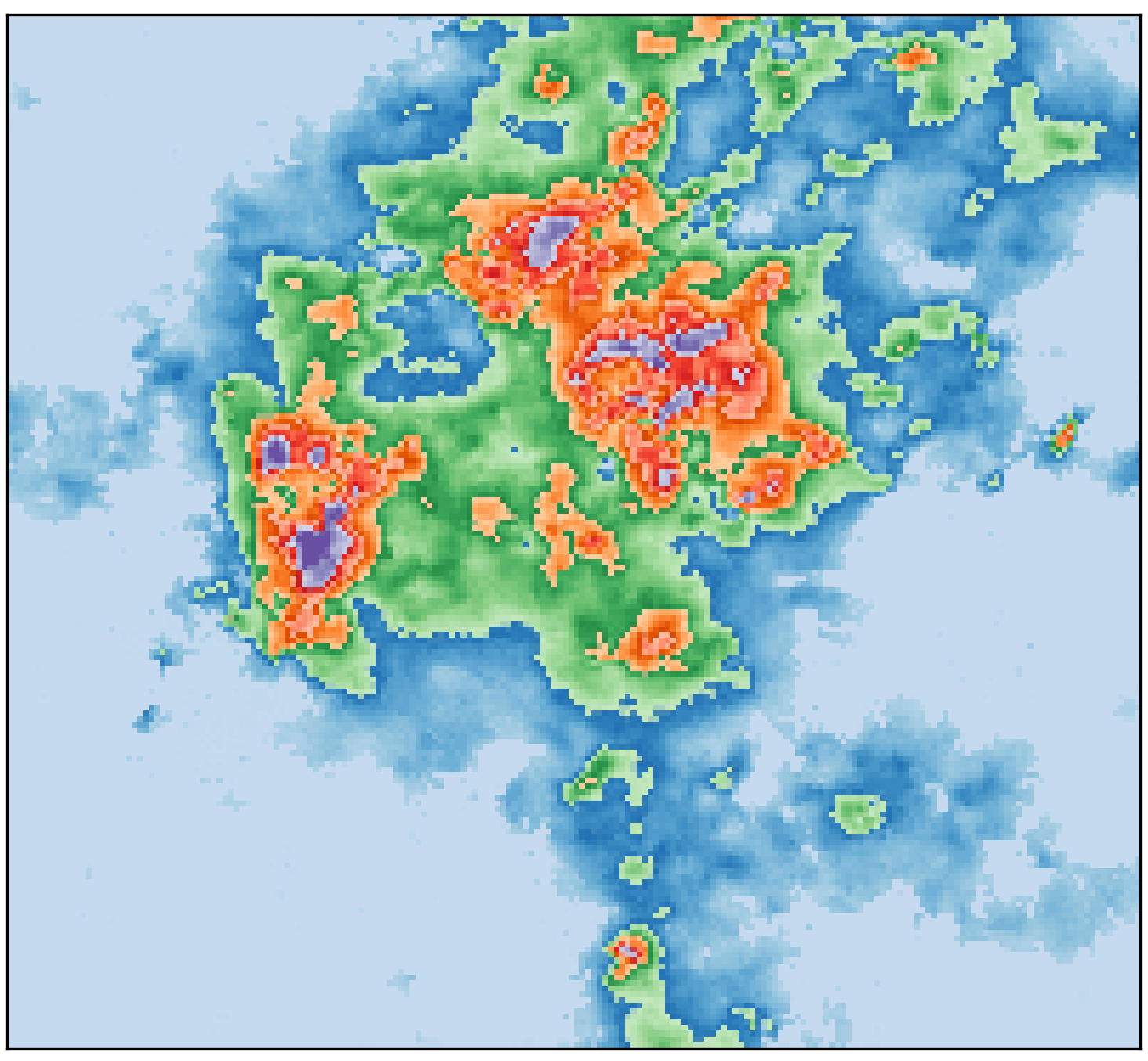} }}
    \subfloat[\centering MRMS]{{\includegraphics[width=0.27\linewidth]{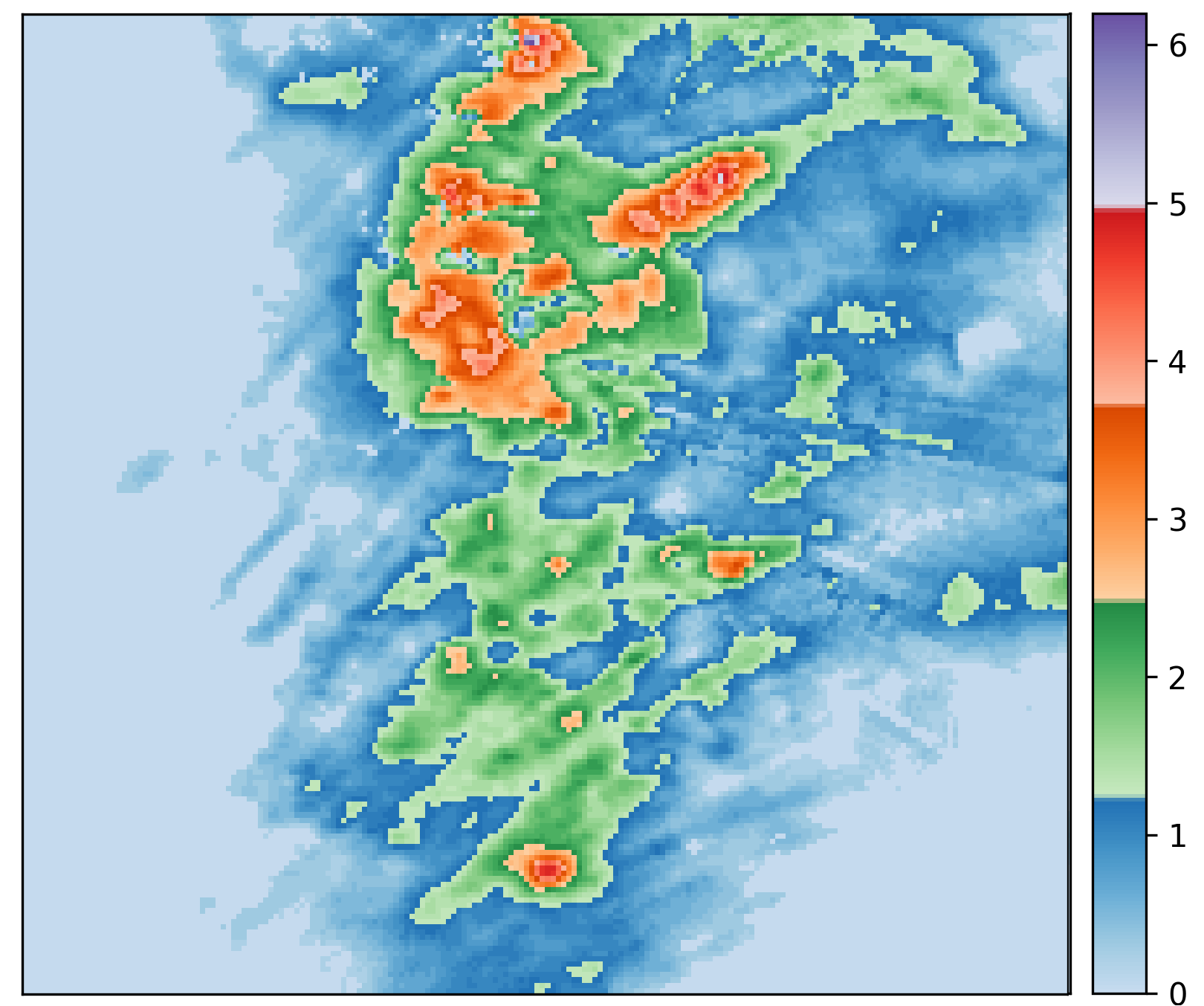} }}
    \caption{Illustration of a unified test for enhancing the satellite precipitations. The IMERG is calibrated by the corrector resulting in (b) corrected IMERG, and (c) is sampled by conditioning on the corrected IMERG. A comparison is made between (c) corrected and downscaled IMERG and (d) MRMS data.}
    \label{fig:snapshot}
\end{figure*}
\begin{figure}
    \centering
    \includegraphics[width=\linewidth]{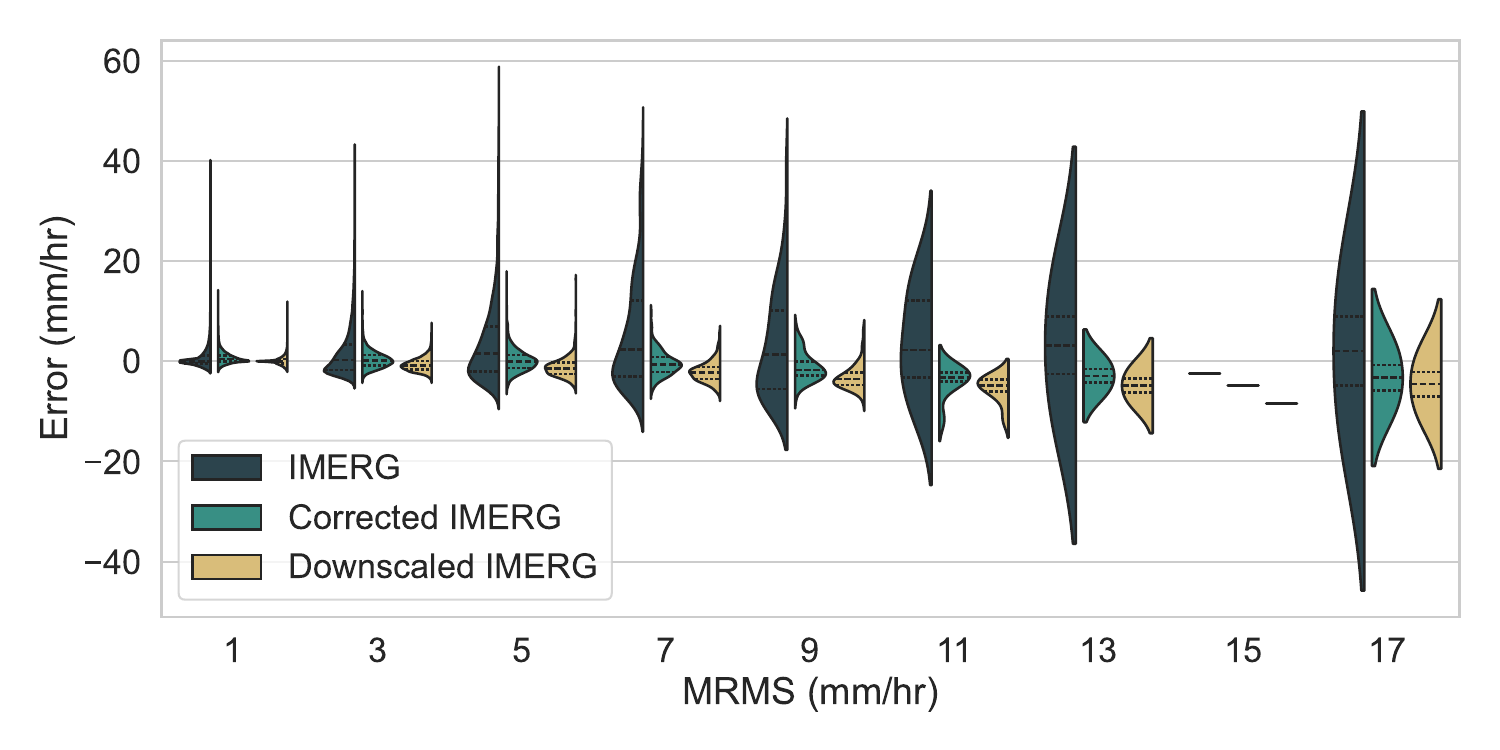}
    \caption{Unified correction and downscaling: The error distribution comparison between IMERG, corrected IMERG and downscaled IMERG observations.}
    \label{fig:vio-two-stage}
\end{figure}

\begin{table*}[t]
    \centering
    \begin{tabular}{c|cccc}
        \hline
        Data Comapred with MRMS & RMSE & CRPS & CC & SSIM \\ \hline
        IMERG & 3.3410 & 1.7264 & 0.5417 & \textbf{0.6344} \\ 
        Corrected IMERG & 1.3597 & 0.9701  & \textbf{0.6615} & 0.5736 \\ 
        Corrected and Dowscaled IMERG &\textbf{ 1.0846} & \textbf{0.7432}  & 0.6426 & 0.5766 \\ 
        \hline
    \end{tabular}
    \caption{Evaluation for Unified task: \textbf{IMERG to MRMS}. RMSE, CRPS, CC, and SSIM are used to evaluate the overall performance and synthesized image quality between original MRMS data across all stages.}
    \label{tab:two-stage}
\end{table*}
\subsection{Downscaling}
To evaluate the downscaling performance independently of distribution shifts, we conducted an experiment to downscale LR MRMS to HR MRMS. This experiment aimed to assess the model's ability to add fine-scale details to LR data. Table \ref{tab:downscaling} demonstrates significant improvements across all metrics for our diffusion model compared to linear regression.

Figure \ref{fig:vio-M2M} compares the error distributions of LR MRMS and downscaled MRMS relative to HR MRMS. To simulate real-world scenarios where precise ground truth values are unavailable at every location, we randomly sampled nearby pixels within the LR pixel area. The results show that our downscaled predictions exhibit errors tightly centered around zero, indicating high accuracy, while LR MRMS shows a wider error distribution.


\subsection{Unified Correction and Downscaling}

To evaluate the combined performance of our framework, we applied both the correction and downscaling models sequentially to downscale IMERG data to the resolution of MRMS observations. Figure \ref{fig:snapshot} illustrates the results, showcasing the detailed cloud-like formations and precipitation patterns captured by our method. Notably, the correction model plays a crucial role in providing a more accurate foundation for downscaling.

Figure \ref{fig:vio-two-stage} compares the error distributions of IMERG, corrected IMERG, and downscaled IMERG relative to MRMS data. Corrected IMERG significantly reduces bias and variation across all precipitation intensities. Downscaled IMERG performs particularly well at lower precipitation values, effectively refining details and capturing subtle variations. However, the increased number of zero-value pixels in the downscaled product leads to some underestimation at higher precipitation intensities.

Table \ref{tab:two-stage} highlights that the downscaled IMERG product exhibits the best performance to MRMS data in terms of RMSE and CRPS, indicating its suitability for both visual interpretation and numerical weather modeling applications. For correlation-foused metrics i.e. CC and SSIM, since the our method is to tweak the LR image with the sampled residual, the overall correlation is lower after the correction task, which is a potential area for further exploration.

\subsection{Out-of-Distribution Evaluation}
\begin{figure}

\subfloat[\centering New York, NY]{{\includegraphics[width=\linewidth]{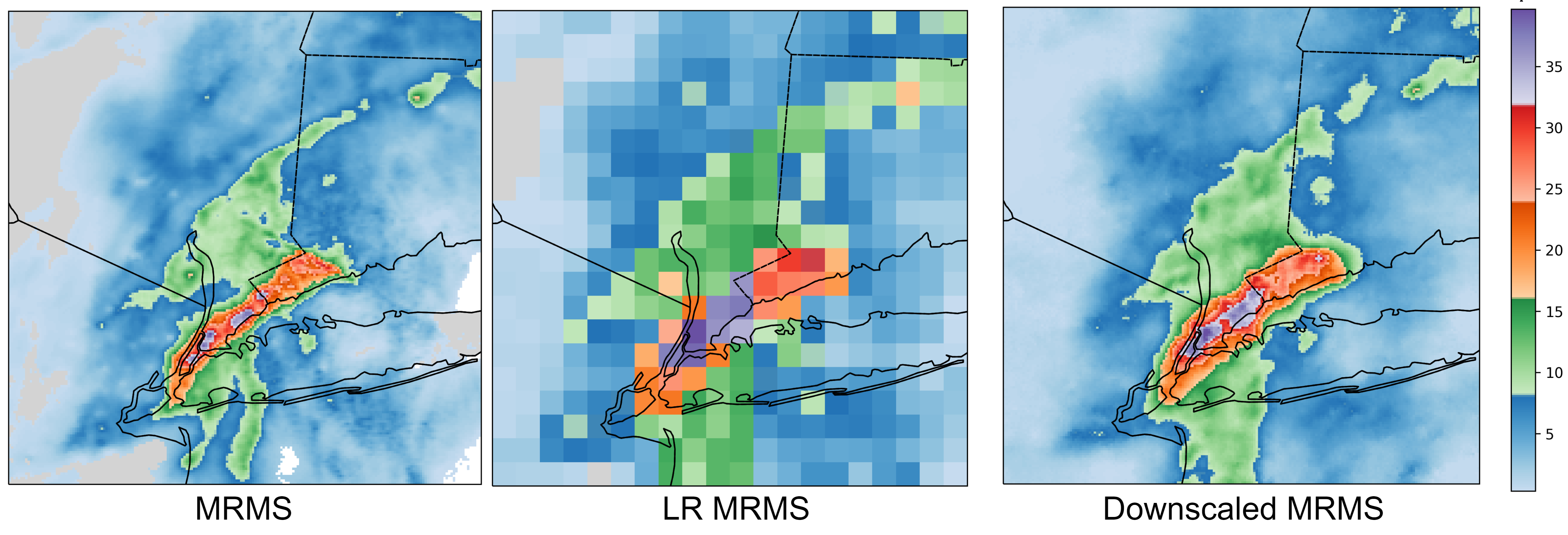} }}\\
    \subfloat[\centering San Jose, CA]{{\includegraphics[width=\linewidth]{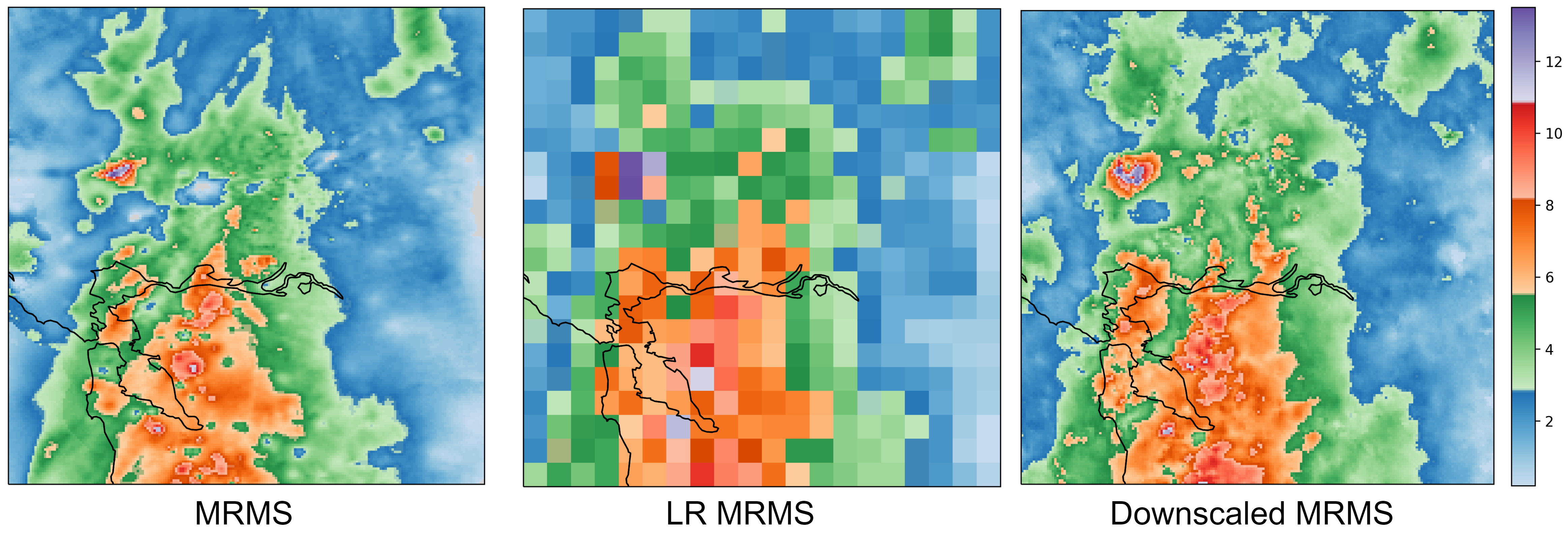} }}\\
    \subfloat[\centering Portland, OR]{{\includegraphics[width=\linewidth]{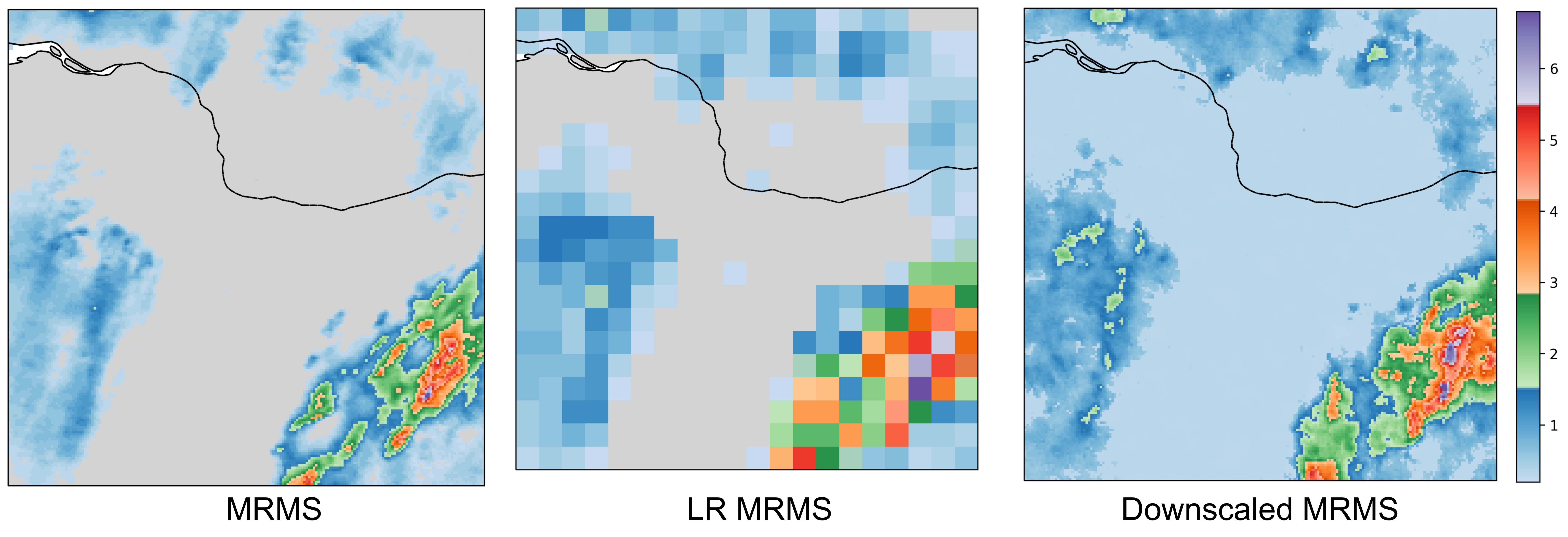} }}\\
    \caption{Downscaling with Seattle-data trained model on other regions}
    \label{fig:other regions}
\end{figure}

\begin{figure}
    \subfloat[\centering New York, NY]{{\includegraphics[width=\linewidth]{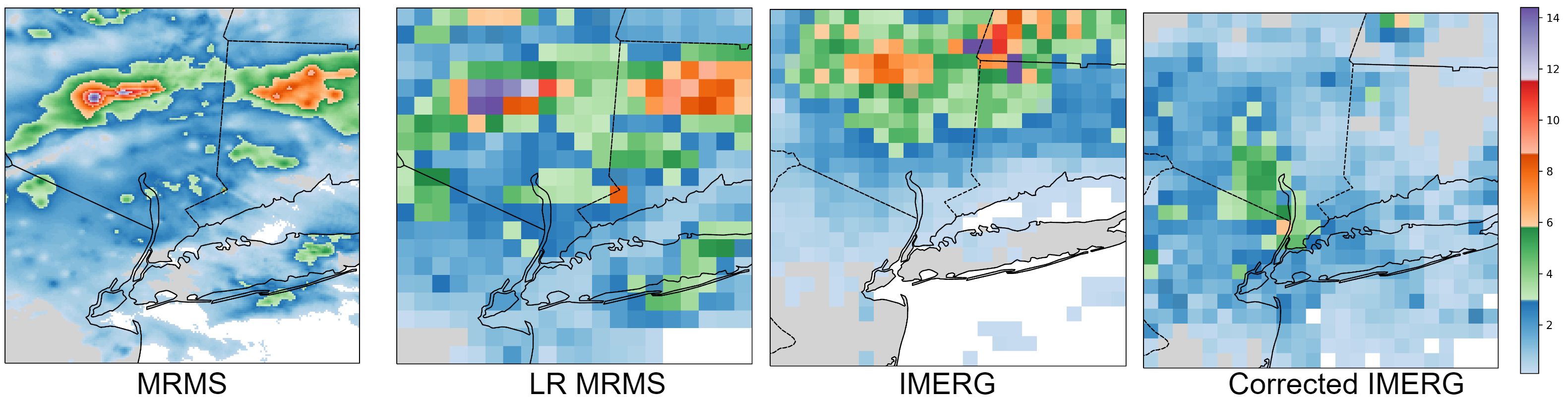} }}\\
    \subfloat[\centering San Jose, CA]{{\includegraphics[width=\linewidth]{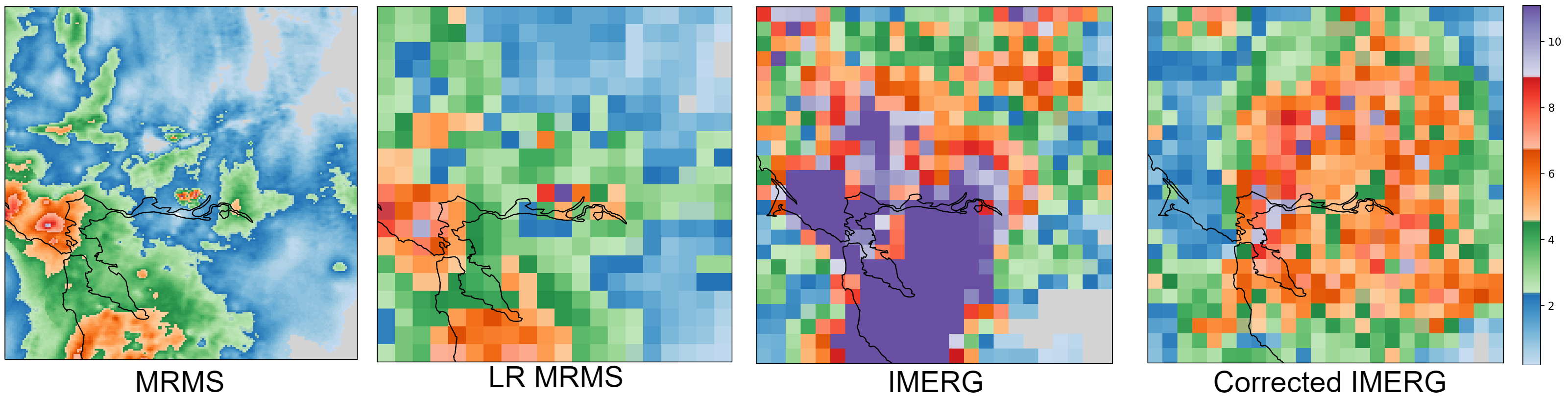} }}\\
    \subfloat[\centering Portland, OR]{{\includegraphics[width=\linewidth]{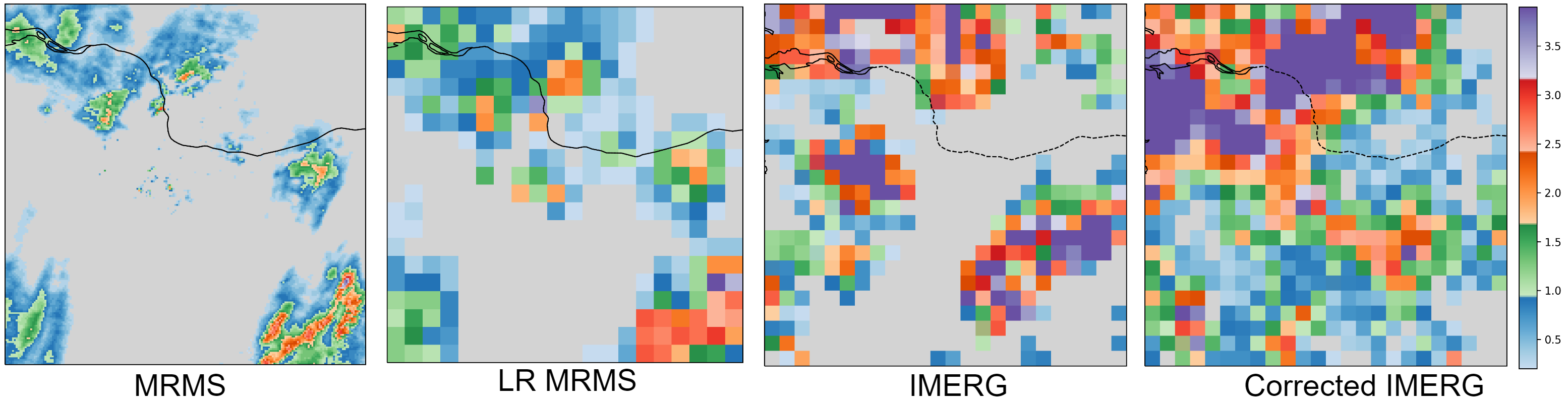} }}\\
    \caption{Correction using a model trained with Seattle data for other regions}
    \label{fig:corrector_other_regions}
\end{figure}

To assess the generalizability of our models, we applied them to data from three additional cities in the United States: New York, Portland, and San Jose (Figures \ref{fig:other regions} and \ref{fig:corrector_other_regions}). The downscaling model effectively reproduced fine-scale details in these new regions, indicating good generalization capabilities. However, the correction model's performance was less consistent, likely due to regional variations in precipitation characteristics and the influence of local rain gauge data assimilation in the IMERG and MRMS products. 

\subsection{Modeling Precipitation with ONLY Precipitation}

One of the primary objectives of this work is to explore the strengths of purely computer vision methods for enhancing precipitation data. Precipitation is a complex variable to work with. This is partly due to the fact that it is sparsely spatially distributed compared to other meteorological variables, like temperature and wind direction. In particular, precipitation is a result of meteorological interactions. Due to such features, it is compelling to see if neural networks can learn the underlying mechanisms. We realize that directly predicting precipitation can be ineffective due to its sparsity. Models tend to underestimate actual values and struggle to preserve the original data due to its stochastic nature. This limitation is particularly evident in our experiments with the correction model which works in low resolution and demonstrates success due to the reduced sparsity.

It is also worth noting that the inherent uncertainty in precipitation distributions poses a challenge for traditional synthesis methods that rely on Gaussian distribution to capture the skewness. To overcome this, we propose learning the residuals between the target and the input, which is based on our finding that the residuals exhibit less skewed distributions compared to the original precipitation values. Consequently, the diffusion model can concentrate on generating fine-scale precipitation details rather than reconstructing the entire sample. However, it is important to note that this approach assumes the availability of a LR data that accurately represents the large-scale statistical properties corresponding to the HR data. In particular, this highlights the critical role of the correction model, which proved to be more influential than originally anticipated, especially when working with operational datasets like IMERG and MRMS. With these datasets, we encountered challenges derived from observational biases to different instruments and various data processing algorithms. These factors introduce regional variations in data distributions, even though both datasets aim to estimate the same variable. Future work will prioritize addressing these distribution shifts more comprehensively.

%% file: 8-conclusion.tex
\section{Conclusion}
This study introduces PrecipDiff, a novel framework leveraging diffusion models and residual learning to address the critical need for high-resolution precipitation data, particularly in regions with limited ground-based monitoring. PrecipDiff effectively tackles two key challenges: correcting biases in satellite precipitation products (SPPs) and downscaling SPP estimates to finer resolutions. Our approach demonstrates substantial improvements in bias reduction and spatial detail, as independent tasks and as a unified framework.

A key innovation of PrecipDiff is its reliance solely on precipitation data, demonstrating the potential of purely data-driven computer vision techniques for enhancing SPPs. This has significant implications for data-scarce regions heavily reliant on satellite observations for applications like disaster preparedness and water resource management. To our knowledge, this is the first study to apply diffusion models exclusively to operational precipitation datasets and for bias correction of SPPs.